# Skin Disease Detection and Classification of Actinic Keratosis and Psoriasis Utilizing Deep Transfer Learning

Fahud Ahmmed[1], Md. Zaheer Raihan[2], Kamnur Nahar[3], D.M. Asadujjaman[4],
Md. Mahfujur Rahman[5], Abdullah Tamim[6]
Dept. of Computer Science & Engineering, Varendra University, Rajshahi [1,2,3,4,5,6]
Dept. of Computer Science & Engineering, Khulna University of Engineering & Technology [4]
Dept. of Computer Science & Engineering, Rajshahi University of Engineering & Technology [5]
Dept. of Computer Science & Engineering, Rajshahi University [6]
fahudahmmed@gmail.com, zaheerraihan09@gmail.com, kamnurnaharshova@gmail.com,
asadujjaman2207557@stud.kuet.ac.bd, mahfujur@vu.edu.bd, tamim.cse.vu@gmail.com

*Abstract*— **Skin diseases can arise from infections, allergies, genetic factors, autoimmune disorders, hormonal imbalances, or environmental triggers such as sun damage and pollution. Skin diseases such as Actinic Keratosis and Psoriasis can be fatal. These are treatable if identified early. However, its diagnostic methods are expensive and not widely accessible. In this study, a novel and efficient method for diagnosing skin diseases using deep learning techniques has been proposed. This approach employs a modified VGG16 Convolutional Neural Network (CNN) model. This model includes several convolutional layers. The VGG16 model has been employed using ImageNet weights and modified top layers. The top layer is modified by fully connected layers and a final softmax activation layer to obtain the result. The dataset analyzed is publicly available and titled "Skin Disease Dataset". The VGG16 architecture does not include augmentation by default; data augmentation is typically performed through rotation, shifting, and zooming during preprocessing prior to model training. The proposed methodology achieved 90.67% accuracy using the modified VGG16 model, demonstrating reliability in classifying skin diseases. The modified pre-trained model showed promising results, increasing its potential for real-world applications.**

## I. INTRODUCTION

Artificial Neural Networks (ANN) have transformed dermatology by offering innovative methods for diagnosing, classifying, and treating skin diseases. Inspired by human brain architecture, ANNs efficiently manage complex data sets and interpret visual information, making them valuable for dermatological diagnoses.

The skin covers the body and serves essential functions, including protection from physical, chemical, and biological threats. Skin diseases are common health issues faced by underprivileged populations in many countries [1]. WHO estimates that approximately 1.8 billion individuals are affected by skin diseases at any given time [2]. Skin diseases may result in skin cancer. Annually, 2 to 3 million non-melanoma skin cancers and 132,000 melanoma skin cancers are diagnosed worldwide [3]. The impact of disorders is influenced by climate, economic conditions, literacy levels, and the cultural and social lifestyles of communities across different geographic areas. Bangladesh experiences a high occurrence due to its humid climate and growing population. Many patients pursue treatment late due to a lack of awareness about the disease. In 2020, skin disease caused 1,131 deaths in Bangladesh, representing 0.16% of total deaths [4], with a prevalence rate of 11.16–63% among the population [5].

Dermatology now incorporates artificial neural networks to analyze various data, including patient history, symptoms, laboratory test results, and crucially, images or scans of skin lesions. ANN can recognize complex patterns and features effectively. The ANN classifies, diagnoses, and predicts skin diseases with high accuracy.

## II. RELATED STUDY

Six techniques—Balanced Random Forest, Balanced Bagging, AdaBoost, Random Forest, Logistic Regression, and Balanced Bagging & SVM—were evaluated using a dataset of 2,453 images. Dermoscopic images of skin diseases, including melanoma, melanocytosis, basal cell carcinoma (BCC), squamous cell carcinoma (SCC), actinic keratosis (AK), seborrheic keratosis (SK), and nevi (moles), have been used to train this model.

A recent study by Van-Dung Hoang et al. found that the EfficientNetB4-CLF model achieved an accuracy of 89.97%, a recall of 86.13%, and a false positive rate of 0.39%. DenseNet 169 outperforms EfficientNetB4 in image classification. ImageNet supplied the pre-trained network to CNNs. CyclicLR was employed to adjust the learning rate periodically. Backpropagation updated the weight values in the CNN architecture. It used 24,530 images resized to 256x192px, allocated as 10% test data, 80% training data, and 10% validation data [6]. Sadia Ghani Malik, Syed Shahryar Jamil, Abdul Aziz, and others developed a robust skin disease detection and classification system using deep neural networks, as detailed in their paper titled "High-Precision Skin Disease Diagnosis through Deep Learning on Dermoscopic Images."

This paper proposes a deep neural network based on a seven-layer CNN architecture. It achieved 87.64% accuracy in tests utilizing the ISIC dataset of dermoscopic images [7]. The DRANet architecture for distinguishing eleven classes of skin conditions is introduced in the paper titled "A Visually Interpretable Deep Learning Framework for Histopathological Image-Based Skin Cancer Diagnosis." The small model size and high classification accuracy make it suitable for real-world applications. They achieved the following results: Accuracy of 86.8% for eleven skin conditions [8]. Zhang et al. aimed to enhance skin disease diagnosis by integrating deep neural networks with human knowledge. The system achieved an accuracy of 87.25%, with a standard deviation of 2-5%, utilizing dermoscopy images. GoogleNet InceptionV3, pre-trained on over 1 million images, was utilized to improve their model [9]. In December 2019, the report titled "Dermatological Classification Using Deep Learning of Skin Images and Patient Background Knowledge" was authored by K. Sriwong, S. Bunrit, K. Kerdprasop, and Nittaya Kerdprasop. This study examines deep learning models utilizing pre-trained networks, recurrent neural networks, and CNNs, applied to image data and patient-specific background knowledge. Their method improved classification accuracy from 79.29% to 80.39% [10].

Our proposed methodology offers enhanced efficiency compared to previously suggested methodologies for deep learning models.

## III. METHODOLOGY

### A. Dataset and Experiment:

We utilized the "Skin Disease Dataset," comprising a total of 2,400 photos across all categories. This dataset is evenly distributed and balanced. The dataset comprises photos from three classes, each containing a total of 800 samples. The dataset has been partitioned into two segments: test data and training data. The test dataset has 150 photos per class, representing 18.8% of the total data for that class, while the training dataset contains 650 images per class, accounting for 81.2% of the total data for that class. Our dataset can be obtained from public resources labeled "Skin Disease Dataset," which has thousands of data entries [11].

Table I illustrates the distribution, while Fig. 1. depicts the ratio of each disease class in our bespoke dataset. The data training process in machine learning is the phase in which a model acquires patterns from the training data. Throughout this procedure, the model is populated with input data attributes and their corresponding output labels. It subsequently readjusts its parameters and weights to minimize the discrepancy between its predictions and the actual labels.

Moreover, the model's learning process necessitates meticulous attention to data quality and parameter optimization to attain optimal outcomes. Our methodology exclusively focuses on the detection and classification of dermatological conditions. Figure 2 displays representative photos from our collection.

### B. Training Model:

*a) Deep Transfer Learning Model (VGG16):* The essential elements of every CNN model comprise convolutional layers, pooling layers, activation functions, fully connected layers, and an output loss layer. Convolutional Neural Networks (CNNs) are distinguished for their outstanding efficacy in visual and natural language tasks. VGG16, created by the Visual Geometry Group at the University of Oxford, is a prominent convolutional neural network architecture. Feature extraction in VGG16 involves employing the network to recognize and specify the most salient characteristics from input images.

The VGG16 architecture exhibits a simple and efficient design, with 16 weight layers, 13 convolutional layers, and 3 fully connected layers.

The classification head has been modified for the proposed VGG16 architecture. The revised head has two interconnected dense layers with 1024 and 512 units, respectively, with a dropout rate of 0.5, succeeded by a final softmax layer with 3 nodes for classification output. The VGG16 was primarily developed to examine the influence of depth on deep neural networks employed for image classification applications.

Here is an overview of the VGG16 architecture:

Input Layer: VGG16 accepts a fixed-size RGB image with dimensions of 150x150 pixels as input.

Convolutional Blocks: It comprises 13 convolutional layers, achieving efficacy using a rectified linear unit (ReLU) operation. In convolutional layers, tiny filters measuring 3x3 pixels utilize a stride of 1, whereas max pooling layers employ 2x2 filters with a stride of 2 for down sampling.

Fully Connected Layers: The subsequent three layers are totally joined. The initial two fully-connected layers consist of 4,096 neurons each. The third consists of 1,000 neurons, corresponding to each of the 1,000 ImageNet classes.

Activation Functions: The Rectified Linear Unit (ReLU) activation function is efficient in convolutional neural networks

Table I
Details of Dataset

| Class | CXR/Class | Data Splitting | |
|---|---|---|---|
| | | *Train* | *Test* |
| Actinic Keratosis | 800 | 650 | 150 |
| Psoriasis | 800 | 650 | 150 |
| Normal | 800 | 650 | 150 |

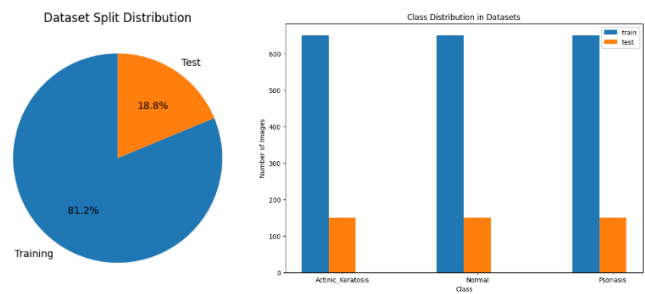

Fig. 1. Dataset Distribution

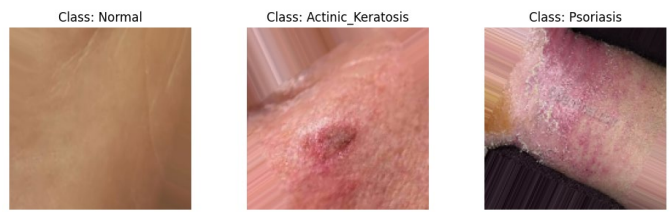

Fig. 2. Sample images of Dataset.

(CNNs) because of its simplicity and computational benefits. ReLU functions are utilized in all convolutional and fully connected layers, with the exception of the final output layer.

$$relu(x) = max(0, x) \quad (1)$$

The final layer employs the SoftMax activation function to provide probabilities for each class in the ImageNet dataset for multi-class detection.

$$softmax(x) = \frac{e^{x_i}}{\Sigma_j e^{x_j}} \quad (2)$$

Loss Functions: Categorical Crossentropy is utilized to assess the discrepancy between the projected class and the actual class. Cross-entropy yields favorable outcomes in instances of inadequate expectations.

$$L(y, \hat{y}) = -\Sigma_{i=1}^{c} y_i \log(\hat{y}) \quad (3)$$

The 3x3 filters with a stride of 1 enable VGG16 to capture subtle spatial information in the input images. Its clarity and solid structure facilitate comprehension and implementation. The convolution technique involves utilizing a 3x3 kernel with adjustable parameters. Nonetheless, VGG16 is computationally demanding, possessing a substantial number of parameters that render its training and implementation resource-intensive in comparison to newer designs such as ResNet and MobileNet.

*C. Hyperparameters:*

Hyperparameters in CNNs encompass input size, activation functions, learning rate, batch size, epochs, optimizer, and dropout rate, all of which significantly influence model learning and performance. The learning rate regulates parameter updates, whereas dropout mitigates overfitting. Table II presents the hyperparameters utilized in the experimental model.

## IV. RESULT AND DISCUSSION

The disease detection rate, commonly referred to as the diagnostic rate, is a crucial metric in the fields of epidemiology and public health. The proposed model has demonstrated a substantial outcome in our analysis. The Receiver Operating Characteristic (ROC) curve, Confusion Matrix, and classification reports, encompassing F1-Score, Precision, and Recall, are calculated herein. Fig. 4 and Fig. 5. exhibit the confusion matrix and the ROC curve, respectively. Table III contains the classification report.

The ROC curve indicates that our model has a minimum Area Under Curve (AUC) value of 0.9496 and a maximum AUC value of 0.9997. The diagonal values from the Confusion Matrix in Fig. 4 indicate the precisely predicted outcomes of our model.

Table IV presents a comparative study of our suggested method and already established methodologies, indicating that our approach yields significant results, surpassing those of certain recent studies in the relevant field.

## V. CONCLUSION

Identifying skin disorders will aid in lowering death rates, preventing disease spread, and mitigating the severity of the condition. Conventional clinical techniques for diagnosing skin disorders entail numerous costly and time-intensive processes. Image processing techniques facilitate the advancement of automated dermatological screening at an early stage. Consequently, feature extraction is crucial for the effective classification of dermatological conditions.

Table III
CLASSIFICATION REPORT OF VGG16

| Class | Precision | Recall | F1-Score | Support |
|---|---|---|---|---|
| Actinic Keratosis | 0.90 | 0.83 | 0.87 | 150 |
| Psoriasis | 1.00 | 0.98 | 0.99 | 150 |
| Normal Skin | 0.83 | 0.91 | 0.87 | 150 |
| Accuracy | - | - | 0.91 | 450 |
| Macro AVG | 0.91 | 0.91 | 0.91 | 450 |
| Weighted AVG | 0.91 | 0.91 | 0.91 | 450 |

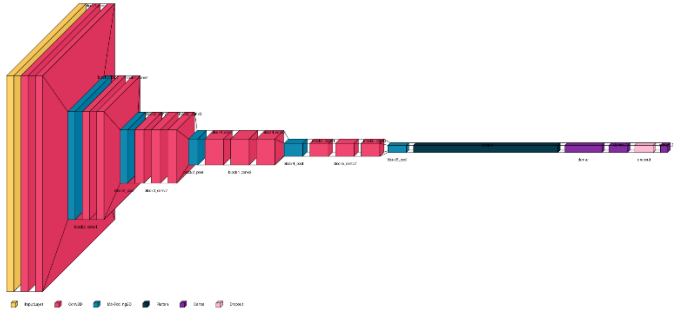

Fig. 3. Modified VGG16 Architecture

Table II
MODEL PARAMETERS FOR INPUT AND CLASSIFICATION STAGE

| Parameters | Approach |
|---|---|
| Input Size | 150 × 150 |
| No. of Epochs | 150 |
| Batch size | 8 |
| Activation | Softmax |
| Optimizer | Adam |
| Learning Rate | 0.0001 |

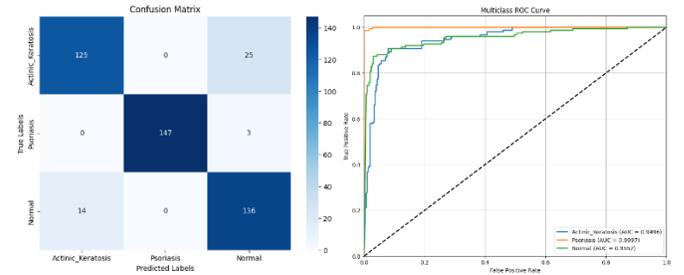

Fig. 4. Confusion Matrix of VGG16     Fig. 5. ROC Curve of VGG16

Table IV
COMPARISON OF THE PROPPOSED MODELS WITH OTHER DEEP LEARNING APPROACHES

| References | No. of Images | | | Total Class | Used Model | F1-Score | Precision | Recall | Accuracy |
|---|---|---|---|---|---|---|---|---|---|
| | Actinic Keratosis | Psoriasis | Normal | | | | | | |
| [8] | - | 872 | 452 | 3 | Inception V3 | - | 0.9422 | 0.9589 | 98.70% |
| | | | | | SE_ResNet101 | - | 0.9661 | 0.9617 | 98.50% |
| | | | | | SE_ResNeXt101-32x4d | - | 0.9492 | 0.9426 | 98.70% |
| | | | | | EfficientNetB4 (modified) | - | 0.9615 | 0.9726 | 98.70% |
| [9] | 850 | - | - | 7 | EfficientNetB4 | - | 0.8900 | 0.8613 | 89.97% |
| | | | | | InceptionV3 | - | 0.8614 | 0.8156 | 86.95% |
| | | | | | ResNet-50 | - | 0.8600 | 0.8061 | 87.61% |
| | | | | | DenseNet169 | - | 0.8714 | 0.8445 | 88.46% |
| [12] | - | - | - | Benign | Custom CNN | 0.84 | 0.91 | 0.79 | 84.71% |
| | | | | Malignant | | 0.82 | 0.78 | 0.87 | |
| | | | | Basal cell carcinoma | | 0.95 | 0.93 | 0.96 | |
| [15] | - | 132 | - | Basal cell carcinoma | ResNet-50 | 0.879 | 0.8824 | 0.8750 | 87.25 ± 2.24% |
| | | | | Melanocytic nevus | | 0.887 | 0.8906 | 0.8837 | |
| | | | | Psoriasis | | 0.885 | 0.8855 | 0.8855 | |
| | | | | Seborrheic keratosis | | 0.797 | 0.7907 | 0.8031 | |
| Proposed Model | 800 | 800 | 800 | Actinic Keratosis | VGG16 (Modified) | 0.87 | 0.90 | 0.87 | 90.67% |
| | | | | Psoriasis | | 0.99 | 1.00 | 0.99 | |
| | | | | Normal Skin | | 0.87 | 0.83 | 0.87 | |

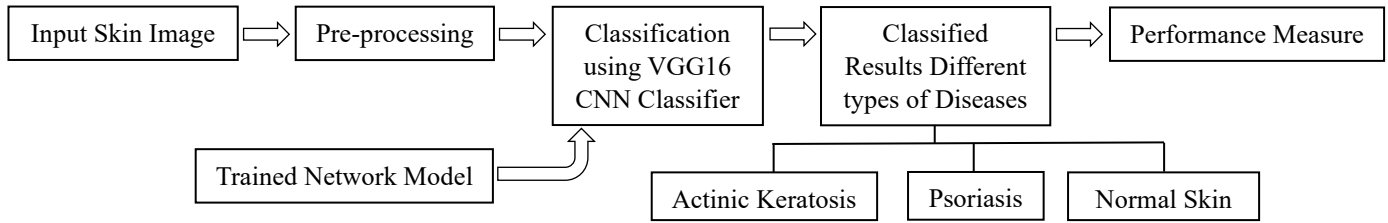

Fig. 6. Block diagram of multiclass classification using the proposed framework

The research utilized a modified pre-trained CNN model with a top layer integrated with SVM for detecting purposes. The improved VGG16 model achieves an accuracy of 90.67%. VGG16 shown exceptional efficacy in skin disease diagnosis through the utilization of deep convolutional neural networks. The model's accuracy is remarkable and indicates significant promise.

It possesses several notable advantages, including elevated accuracy, benefits of transfer learning, and interpretability. Addressing dataset limitations, ensuring strong generalization, and fostering collaboration between medical and machine learning professionals are essential for integrating VGG16 into clinical workflows.